\documentclass[preprint,12pt]{elsarticle}

\usepackage{amsmath,amssymb,bm,mathrsfs}
\usepackage{amsthm}
\usepackage{booktabs}
\usepackage{graphicx}
\usepackage{multirow}
\usepackage{float}
\usepackage{algorithm}
\usepackage{algorithmic}
\usepackage[protrusion=true,expansion=false]{microtype}
\usepackage{url}
\usepackage{xcolor}
\usepackage{placeins}
\usepackage[a4paper,margin=1in]{geometry}
\usepackage[colorlinks=true,linkcolor=black,citecolor=black,urlcolor=blue,bookmarks=true,bookmarksnumbered=true]{hyperref}
\hypersetup{pdftitle={HTNN-MSC: Multi-view Subspace Clustering with a Hyperbolic Tangent Tensor Low-Rank Surrogate},pdfauthor={Yiwen Dou, Yiqiang Ding},pdfsubject={Multi-view subspace clustering},pdfkeywords={multi-view clustering, low-rank tensor, nonconvex surrogate, ADMM}}

\biboptions{numbers,sort&compress}
\journal{Pattern Recognition}

\newtheorem{proposition}{Proposition}
\newtheorem{theorem}{Theorem}

\newcommand{\R}{\mathbb{R}}

\newcommand{\cZ}{\mathcal{Z}}
\newcommand{\cG}{\mathcal{G}}
\newcommand{\cB}{\mathcal{B}}
\newcommand{\cW}{\mathcal{W}}

\newcommand{\PhiT}{\Phi_{\theta}}
\newcommand{\norm}[1]{\left\lVert #1\right\rVert}
\newcommand{\inner}[2]{\left\langle #1,#2\right\rangle}
\DeclareMathOperator{\rank}{rank}

\DeclareMathOperator{\sech}{sech}

\DeclareMathOperator{\dist}{dist}

\begin{document}

\begin{frontmatter}
\title{HTNN-MSC: Multi-view Subspace Clustering with a Hyperbolic Tangent Tensor Low-Rank Surrogate}
\author[inst1]{Yiwen Dou}
\ead{douyw@ahpu.edu.cn}
\author[inst1]{Yiqiang Ding}
\ead{2250930101@stu.ahpu.edu.cn}
\affiliation[inst1]{organization={School of Computer and Information, Anhui Polytechnic University},city={Wuhu},country={China}}

\begin{abstract}
High-order dependencies among views can be modeled by tensor singular value decomposition (t-SVD) in multi-view subspace clustering. However, the tensor nuclear norm often used with this framework assigns the same asymptotic marginal penalty to singular values of different magnitudes, which may excessively shrink dominant components that encode cluster structure. We propose HTNN-MSC, which replaces this norm with a bounded hyperbolic tangent tensor rank surrogate. The surrogate is strictly concave on the positive half-axis and has a Lipschitz continuous derivative. Its marginal penalty decreases as singular values grow. We develop an alternating direction method of multipliers solver in which the tensor subproblem decomposes into scalar frequency-domain shrinkage problems. Safeguarded multi-start Newton iterations and candidate-objective comparison implement the resulting hyperbolic tangent thresholding operator. The shape parameter is scaled by the input spectrum. We also establish o-minimal definability of the augmented objective. Under explicit boundedness, descent, relative-error, and exact-block conditions, the Kurdyka--{\L}ojasiewicz framework guarantees convergence of the full sequence to a KKT point. On 23 datasets, HTNN-MSC achieves a mean clustering accuracy of 0.8404 and an average accuracy rank of 2.78 against nine representative methods. Friedman tests and Holm-adjusted Wilcoxon comparisons reveal significant overall differences, although several pairwise differences remain nonsignificant. HTNN-MSC records 19 wins, one tie, and three losses against the convex t-SVD-MSC skeleton. A same-skeleton ablation also shows that the proposed surrogate does not uniformly outperform alternative nonconvex penalties.
\end{abstract}

\begin{keyword}
multi-view clustering \sep subspace clustering \sep low-rank tensor \sep nonconvex surrogate \sep tensor singular value decomposition \sep ADMM
\end{keyword}
\normalsize
\end{frontmatter}

\section{Introduction}
\label{sec:introduction}

Multiple sensors, modalities, or feature extractors can provide complementary descriptions of a common set of objects. For example, intensity, texture, and shape features offer distinct characterizations of an image, whereas multilingual versions of a news article provide different textual views. By exploiting information shared across these representations together with view-specific cues, multi-view clustering seeks to partition data without labels~\cite{liu2026survey}. Multi-view subspace clustering (MVSC) addresses data whose samples approximately occupy a collection of low-dimensional subspaces. It learns within-view coefficients that express each sample through other samples, and then converts these coefficients into an affinity graph.

To obtain these coefficients, sparse subspace clustering uses a sparsity prior, whereas low-rank representation adopts a low-rank prior~\cite{elhamifar2013sparse,liu2013robust}. Learning each view independently, however, can miss dependencies between views. Tensor-based MVSC instead stacks the view-specific representations into a third-order tensor and constrains their joint structure. The choice of tensor rank surrogate then determines how the model suppresses noise and retains principal cluster structure.

Many t-SVD-based models use the tensor nuclear norm (TNN)~\cite{lu2020tnn}. Its convexity simplifies optimization, but its derivative with respect to each positive singular value is constant. Small coefficients associated with noise and large coefficients associated with dominant structure therefore receive the same asymptotic marginal penalty. Nonconvex surrogates reduce this penalty as the singular value increases. Logarithmic Schatten-$p$ and generalized nonconvex tensor penalties illustrate this idea in MVSC~\cite{guo2023tlspnm,chen2021gnlta}. Arctangent and refined-sigmoid penalties add bounded alternatives~\cite{sun2023arctangent,gu2024sigmoid}. The remaining question is how a particular penalty affects shrinkage, numerical solution, parameter scaling, and convergence analysis.

We study $\phi_\theta(\sigma)=\tanh(\sigma/\theta)$ as a tensor rank surrogate. This function is bounded and smooth, strictly concave for $\sigma>0$, and finite in derivative at the origin. It shrinks small coefficients while its marginal penalty vanishes for sufficiently large ones. We incorporate this penalty into a self-expression MVSC model and derive an alternating direction method of multipliers (ADMM) solver. Fourier transformation and singular value decomposition reduce the tensor update to independent scalar problems. The resulting mapping, called hyperbolic tangent thresholding (HTT), is evaluated from three starting points with safeguarded Newton updates and objective-value comparison.

The paper makes four contributions.
\begin{enumerate}
    \item We formulate the hyperbolic tangent tensor surrogate and establish its boundedness, strict concavity, Lipschitz derivative, rank-approximation limit, and definability in an o-minimal structure.
    \item We derive a two-penalty ADMM algorithm that matches the implementation. A data-adaptive rule sets $\theta$ from the initial frequency-domain spectrum, while the error-weight parameter remains dataset dependent.
    \item We state the exact-block, descent, and relative-error conditions required for KL-based whole-sequence convergence. The statement distinguishes the theoretical algorithm from the practical multi-start HTT solver and does not treat tanh as semi-algebraic.
    \item We compare ten methods on 23 datasets. HTNN-MSC achieves a mean ACC of 0.8404 and a mean NMI of 0.8029, with respective average ranks of 2.78 and 3.02. Friedman and Holm-adjusted Wilcoxon tests assess cross-dataset differences. Same-skeleton comparisons measure the change from convex soft thresholding and expose cases where other nonconvex surrogates perform better. Repeated post-processing, sensitivity analyses, convergence plots, and runtime measurements describe the numerical behavior of the method.
\end{enumerate}

\section{Related work}
\label{sec:related}

\subsection{Multi-view subspace clustering}

Self-expression reconstructs a sample from other samples in the same subspace. Given $X\in\R^{d\times n}$, the model learns $Z$ from $X=XZ+E$ and applies spectral clustering to an affinity derived from $Z$. Multi-view extensions learn $\{Z_v\}_{v=1}^{V}$ and reconcile view-specific reconstruction with a shared cluster structure.

LT-MSC stacks these matrices and constrains the Tucker multi-rank of the resulting tensor~\cite{zhang2015ltmsc}. t-SVD-MSC rotates the representation tensor and applies a tensor nuclear norm after Fourier transformation~\cite{xie2018tsvd}. ETLMSC learns an essential tensor for spectral clustering~\cite{wu2019essential}. UGLTL jointly learns graphs and a low-rank tensor, whereas WTNNM introduces a weighted tensor nuclear norm into graph-based MVSC~\cite{wu2020ugltl,gao2020wtnnm}. LRTGL and LTBPL combine tensor modeling with graph learning and proximity learning, respectively~\cite{chen2022lrtgl,chen2023ltbpl}.

Full-representation methods store an $n\times n$ matrix for each view. Anchor-based models reduce this cost. LMVSC has linear complexity in the sample number under fixed anchor settings~\cite{kang2020lmvsc}. ESTMC combines anchor representations with nonconvex tensor rank learning~\cite{ji2025estmc}, and TLRLF4MVC adds a tensor low-frequency operator to scalable multi-view clustering~\cite{long2025tlrlf}. GFT-TMVC replaces the fixed sample-mode Fourier basis with a graph Fourier basis to reduce sensitivity to sample ordering~\cite{ji2026gft}. These methods change the representation scale or transform basis. Our model keeps the full representation and changes the singular-value penalty.

\subsection{Nonconvex low-rank regularization}

Rank is measured by the number of nonzero singular values; by contrast, the sum of their magnitudes defines the nuclear norm. MCP and SCAD reduce estimation bias by making the penalty derivative depend on coefficient size~\cite{zhang2010mcp,fan2001scad}. Logarithmic and Schatten-$p$ penalties also reduce pressure on large coefficients, but their derivatives can be unbounded or singular near the origin.

TLSpNM uses a logarithmic Schatten-$p$ penalty, and GNLTA considers a broader family of nonconvex tensor approximations. Arctangent and refined-sigmoid surrogates supply bounded alternatives. HTNN-MSC instead uses tanh, scales it from the input spectrum, and compares it with convex and nonconvex penalties in the same optimization skeleton. Its o-minimal formulation states the conditions needed for a KL convergence argument. The experiments evaluate these properties without assuming that they imply higher accuracy on every dataset.

\section{Notation and preliminaries}
\label{sec:preliminaries}

Matrices are written in uppercase and third-order tensors in calligraphic letters. For view $v$, $X_v=[x_{v,1},\ldots,x_{v,n}]\in\R^{d_v\times n}$ contains $n$ samples, $Z_v\in\R^{n\times n}$ stores self-expression coefficients, and $E_v$ contains reconstruction errors. The numbers of views and clusters are $V$ and $K$. The Frobenius norm is denoted by $\norm{\cdot}_F$, and the matrix inner product is $\inner{A}{B}=\mathrm{tr}(A^\top B)$. Column sparsity is promoted by $\norm{E}_{2,1}=\sum_{j=1}^{n}\norm{E_{:,j}}_2$.

The representation tensor $\cZ\in\R^{n\times n\times V}$ has frontal slices $Z_v$. A cyclic dimension shift rotates it to $\cZ^r\in\R^{V\times n\times n}$. The discrete Fourier transform acts on the last, sample-indexed mode. We write $\widehat{\cZ}=\mathrm{fft}(\cZ^r,[],3)$ and denote its $j$th frontal slice by $\widehat{Z}^{(j)}$. The Fourier mode therefore has length $n_3=n$, rather than $V$. The t-product and t-SVD frameworks behind this construction appear in~\cite{kilmer2011factorization,kilmer2013thirdorder}.

For a real tensor under the unnormalized FFT, $\norm{\cZ}_F^2=n_3^{-1}\norm{\widehat{\cZ}}_F^2$. Conjugate symmetry allows us to process only the first $\lfloor n_3/2\rfloor+1$ Fourier slices and reconstruct the remainder. We define the scaled nonconvex penalty
\begin{equation}
    \PhiT(\cG)=\frac{1}{n_3}\sum_{j=1}^{n_3}\sum_{i=1}^{r_j}
    \phi_\theta\!\left(\sigma_i(\widehat{G}^{(j)})\right),
    \qquad r_j=\rank(\widehat{G}^{(j)}),
    \label{eq:tensor_surrogate_prelim}
\end{equation}
where zero singular values contribute zero. The factor $1/n_3$ matches the proximal derivation to the unnormalized FFT in the implementation.

After optimization, the affinity matrix is
\begin{equation}
    S=\sum_{v=1}^{V}\left(|Z_v|+|Z_v^\top|\right),
    \label{eq:affinity_prelim}
\end{equation}
where absolute values are elementwise. Spectral clustering partitions $S$ into $K$ groups~\cite{ng2002spectral,vonluxburg2007tutorial}.

\section{The proposed HTNN-MSC model}
\label{sec:method}

\subsection{Hyperbolic tangent rank surrogate}

For $\sigma\geq 0$ and $\theta>0$, define
\begin{equation}
    \phi_\theta(\sigma)=\tanh\left(\frac{\sigma}{\theta}\right)
    =\frac{e^{2\sigma/\theta}-1}{e^{2\sigma/\theta}+1}.
    \label{eq:tanh_surrogate}
\end{equation}
Applying $\phi_\theta$ to each Fourier-slice singular value in Eq.~\eqref{eq:tensor_surrogate_prelim} gives the tensor penalty $\PhiT$.

\begin{proposition}[Properties of $\phi_\theta$]
\label{prop:properties}
The following properties hold for $\sigma\geq0$.
\begin{enumerate}
    \item \textbf{Boundedness and normalization.} $\phi_\theta(0)=0$ and $0\leq\phi_\theta(\sigma)<1$.
    \item \textbf{Strict concavity.} For $\sigma>0$,
    \begin{equation}
        \phi_\theta''(\sigma)=-\frac{2}{\theta^2}\sech^2(\sigma/\theta)\tanh(\sigma/\theta)<0.
    \end{equation}
    \item \textbf{Finite and Lipschitz derivative.} $\phi_\theta'(0)=1/\theta$, $\lim_{\sigma\to\infty}\phi_\theta'(\sigma)=0$, and
    \begin{equation}
        |\phi_\theta''(\sigma)|\leq \frac{4}{3\sqrt{3}\theta^2}.
    \end{equation}
    Hence $\phi_\theta'$ is Lipschitz with constant $4/(3\sqrt{3}\theta^2)$.
    \item \textbf{Rank limit.} $\lim_{\theta\to0^+}\phi_\theta(\sigma)$ equals zero at $\sigma=0$ and one for every fixed $\sigma>0$.
    \item \textbf{Definability.} $\phi_\theta$ is definable in the o-minimal structure $\R_{\exp}$.
\end{enumerate}
\end{proposition}

Direct calculus gives the first four properties. The curvature bound follows because $\sech^2(x)\tanh(x)$ has maximum absolute value $2/(3\sqrt3)$ at $\tanh(x)=1/\sqrt3$. Eq.~\eqref{eq:tanh_surrogate} and Wilkie's theorem for the real exponential field give definability~\cite{wilkie1996model}. Tanh is not semi-algebraic. The KL analysis must therefore use the o-minimal structure $\R_{\exp}$.

Shrinkage is governed by $\phi_\theta'(\sigma)=\theta^{-1}\sech^2(\sigma/\theta)$. This derivative is finite and positive near zero, then decays rapidly as the singular value grows. A nuclear-norm penalty instead has a constant derivative for positive singular values. Tanh thus reduces asymptotic shrinkage of dominant spectral components, although this bias-reduction mechanism does not guarantee higher accuracy on every dataset.

\subsection{Optimization model}

We combine view-specific reconstruction errors with the nonconvex tensor penalty.
\begin{equation}
\begin{aligned}
    \min_{\{Z_v,E_v\},\cG}\quad
    &\PhiT(\cG)+\lambda\sum_{v=1}^{V}\norm{E_v}_{2,1}\\
    \mathrm{s.t.}\quad
    &X_v=X_vZ_v+E_v,\quad v=1,\ldots,V,\\
    &\cZ=\cG .
\end{aligned}
\label{eq:model}
\end{equation}
The parameter $\lambda>0$ weights the error term. The auxiliary tensor $\cG$ separates the low-rank penalty from the reconstruction constraints. The algorithm and experiments use this same parameterization.

The affinity construction is identical to that of the convex t-SVD-MSC skeleton. Replacing $\phi_\theta$ with the identity penalty therefore changes singular-value shrinkage without changing graph fusion or spectral post-processing. The ablation study uses this controlled substitution.

\subsection{Data-adaptive surrogate scale}

Datasets can have singular values on very different scales, so a fixed $\theta$ is difficult to choose. We define $\bar\sigma$ as the average of the singular values in the first $\lfloor n_3/2\rfloor+1$ Fourier slices of the input to the first $\cG$-subproblem, and set $\theta=c\bar\sigma$. The main experiments use $c=1$. Formally,
\begin{equation}
    \bar\sigma=\frac{1}{M}\sum_{j=1}^{\lfloor n_3/2\rfloor+1}
    \sum_{i=1}^{\min(V,n)}\sigma_i(\widehat{B}^{(1,j)}),
    \qquad \theta=c\bar\sigma,
    \label{eq:adaptive_theta}
\end{equation}
Here, $M$ denotes the total number of singular values included in the average, and $\cB^{(1)}=\cZ^{(0)}+\cW^{(0)}/\rho_0$. This rule fixes the scale of $\theta$ from the data, but $\lambda$ still requires selection. The best $\lambda$ varies substantially between text and image data.

\section{Optimization}
\label{sec:optimization}

\subsection{Augmented Lagrangian}

Let $r_v=X_v-X_vZ_v-E_v$. Using separate penalty parameters $\mu$ and $\rho$, the augmented Lagrangian is
\begin{equation}
\begin{aligned}
    \mathcal L_{\mu,\rho}={}&\PhiT(\cG)
    +\lambda\sum_v\norm{E_v}_{2,1}
    +\sum_v\inner{Y_v}{r_v}
    +\frac{\mu}{2}\sum_v\norm{r_v}_F^2\\
    &+\sum_v\inner{W_v}{Z_v-G_v}
    +\frac{\rho}{2}\sum_v\norm{Z_v-G_v}_F^2 .
\end{aligned}
\label{eq:augmented}
\end{equation}
Reconstruction and tensor consistency operate at different scales, so the model uses two penalty parameters. With the sign convention $\inner{W_v}{Z_v-G_v}$, the $Z$-equation contains $-W_v/\rho$.

\subsection{View-specific updates}

For fixed $E_v,G_v,Y_v,W_v$, the $Z_v$-subproblem is quadratic. Its minimizer is
\begin{equation}
\begin{aligned}
    Z_v^{k+1}={}&\left(I+\frac{\mu_k}{\rho_k}X_v^\top X_v\right)^{-1}
    \left[G_v^k+\frac{1}{\rho_k}\left(X_v^\top Y_v^k\right.\right.\\
    &\left.\left.+\mu_kX_v^\top X_v-\mu_kX_v^\top E_v^k-W_v^k\right)\right].
\end{aligned}
\label{eq:z_update}
\end{equation}
The matrix $I+(\mu_k/\rho_k)X_v^\top X_v$ is positive definite. Since the penalty ratio changes between iterations, the implementation recomputes the linear solve rather than assuming a single fixed inverse.

After updating $Z_v$, define
\begin{equation}
    T_v=X_v-X_vZ_v^{k+1}+\frac{Y_v^k}{\mu_k}.
\end{equation}
The $E_v$-subproblem separates by columns.
\begin{equation}
    E_{v,:,j}^{k+1}=\max\left(1-\frac{\lambda}{\mu_k\norm{T_{v,:,j}}_2},0\right)T_{v,:,j},
    \label{eq:e_update}
\end{equation}
Here $E_{v,:,j}^{k+1}=0$ when $\norm{T_{v,:,j}}_2=0$. The reconstruction multiplier is then updated by
\begin{equation}
    Y_v^{k+1}=Y_v^k+\mu_k\left(X_v-X_vZ_v^{k+1}-E_v^{k+1}\right).
    \label{eq:y_update}
\end{equation}

\subsection{Hyperbolic tangent tensor thresholding}

The $\cG$-subproblem is
\begin{equation}
    \min_{\cG}\ \PhiT(\cG)+\frac{\rho_k}{2}
    \norm{\cG-\cB^{k+1}}_F^2,
    \qquad \cB^{k+1}=\cZ^{k+1}+\frac{\cW^k}{\rho_k}.
    \label{eq:g_problem}
\end{equation}

\begin{proposition}[Frequency-domain separability]
\label{prop:separability}
Let $t=\sigma_i(\widehat{B}^{(k+1,j)})$. With the scaled surrogate in Eq.~\eqref{eq:tensor_surrogate_prelim} and the unnormalized FFT, Eq.~\eqref{eq:g_problem} separates into scalar problems
\begin{equation}
    \min_{s\geq0}\ f_{t,\tau}(s)
    =\frac12(s-t)^2+\tau\tanh(s/\theta),
    \qquad \tau=\frac1{\rho_k}.
    \label{eq:scalar_htt}
\end{equation}
The matrix solution is obtained by replacing each singular value with the selected minimizer and retaining the corresponding singular vectors.
\end{proposition}

\begin{proof}
Parseval's identity gives $\norm{\cG-\cB}_F^2=n_3^{-1}\norm{\widehat{\cG}-\widehat{\cB}}_F^2$. Multiplying Eq.~\eqref{eq:g_problem} by $n_3$ removes the scale factors in both terms. Each Fourier-slice problem is then a unitarily invariant matrix proximal problem. Its SVD reduces the matrix problem to Eq.~\eqref{eq:scalar_htt}, and conjugate symmetry reconstructs a real tensor.
\end{proof}

For $t\geq0$, any interior stationary point satisfies
\begin{equation}
    h(s)=s-t+\frac{\tau}{\theta}\sech^2(s/\theta)=0,
    \qquad 0<s\leq t.
    \label{eq:stationary}
\end{equation}
Moreover,
\begin{equation}
    h'(s)=1-\frac{2\tau}{\theta^2}\sech^2(s/\theta)\tanh(s/\theta).
\end{equation}
Since $\sech^2(x)\tanh(x)$ increases and then decreases, $h'$ has at most two zeros and $h$ has at most three roots on $(0,t]$. A certified global solver can bracket these monotone intervals and compare all roots with $s=0$.

The experimental implementation uses a less expensive numerical procedure. It starts safeguarded Newton iterations from $\max(t-\tau/\theta,0)$, $t/2$, and $\min(\max(\tau/\theta,10^{-3}t),t)$. Steps are projected onto $[0,t]$, with a half-interval backtracking step when the Newton denominator is too small or nonfinite. After at most 60 iterations per start, all returned candidates and $s=0$ are compared by $f_{t,\tau}$. We call this mapping HTT. The three starts provide candidate solutions but do not certify exhaustive root enumeration.

After shrinking all processed Fourier slices, the inverse FFT and inverse dimension shift produce $\cG^{k+1}$. The multiplier update is
\begin{equation}
    \cW^{k+1}=\cW^k+\rho_k(\cZ^{k+1}-\cG^{k+1}).
    \label{eq:w_update}
\end{equation}
Both penalties double at each iteration until they reach prescribed maxima. Algorithm~\ref{alg:htnn} summarizes the procedure.

\begin{algorithm}[t]
\caption{HTNN-MSC}
\label{alg:htnn}
\begin{algorithmic}[1]
\REQUIRE Views $\{X_v\}_{v=1}^V$, cluster number $K$, $\lambda>0$, scale factor $c>0$.
\STATE Normalize each view and initialize $Z_v^0=G_v^0=W_v^0=0$, $E_v^0=Y_v^0=0$.
\STATE Initialize $\mu_0=10^{-4}$, $\rho_0=10^{-4}$, and cached $X_v^\top X_v$.
\FOR{$k=0,1,\ldots$ until convergence}
    \FOR{$v=1,\ldots,V$}
        \STATE Update $Z_v^{k+1}$ by Eq.~\eqref{eq:z_update}.
        \STATE Update $E_v^{k+1}$ by Eq.~\eqref{eq:e_update}.
        \STATE Update $Y_v^{k+1}$ by Eq.~\eqref{eq:y_update}.
    \ENDFOR
    \STATE Form $\cB^{k+1}=\cZ^{k+1}+\cW^k/\rho_k$.
    \IF{$k=0$ and adaptive scaling is enabled}
        \STATE Compute $\bar\sigma$ and set $\theta=c\bar\sigma$ by Eq.~\eqref{eq:adaptive_theta}.
    \ENDIF
    \STATE Apply FFT to the rotated tensor and compute slice-wise SVDs.
    \STATE Apply HTT to each singular value with $\tau=1/\rho_k$.
    \STATE Reconstruct $\cG^{k+1}$ using conjugate symmetry and inverse FFT.
    \STATE Update $\cW^{k+1}$ by Eq.~\eqref{eq:w_update}.
    \STATE Increase $\mu_{k+1}=\min(2\mu_k,\mu_{\max})$ and $\rho_{k+1}=\min(2\rho_k,\rho_{\max})$.
    \STATE Stop when both $\max_v\norm{X_v-X_vZ_v-E_v}_\infty$ and $\max_v\norm{Z_v-G_v}_\infty$ are at most $10^{-7}$, or at the iteration cap.
\ENDFOR
\STATE Construct $S$ by Eq.~\eqref{eq:affinity_prelim} and apply spectral clustering.
\ENSURE Cluster assignments, affinity $S$, and representations $\{Z_v\}$.
\end{algorithmic}
\end{algorithm}
\FloatBarrier

\section{Convergence conditions and computational complexity}
\label{sec:analysis}

\subsection{KL-based convergence statement}

The KL property converts sufficient descent and a relative-error bound into a finite-length iterate sequence for nonconvex splitting methods. Tanh is not semi-algebraic, but $\phi_\theta$ is definable in $\R_{\exp}$. The quadratic terms, linear constraints, and $\ell_{2,1}$ norm are semi-algebraic. Their finite compositions and sums with $\phi_\theta$ preserve definability in $\mathcal L_{\mu,\rho}$. The augmented objective therefore satisfies the KL property on bounded sets~\cite{bolte2007lojasiewicz,kurdyka1998gradients}.

\begin{theorem}[Conditional whole-sequence convergence]
\label{thm:convergence}
Let $x^k$ collect all primal and dual variables. Suppose the following conditions hold.
\begin{enumerate}
    \item Every $Z$-, $E$-, and $\cG$-subproblem is globally minimized.
    \item Boundedness of the sequence is assumed, together with a proper lower-bounded potential $\Psi$ and a constant $a>0$ satisfying
    $\Psi(x^k)-\Psi(x^{k+1})\geq a\norm{x^{k+1}-x^k}^2$.
    \item A constant $b>0$ and a subgradient $d^{k+1}\in\partial\Psi(x^{k+1})$ satisfy
    \begin{equation*}
        \norm{d^{k+1}}\leq b\norm{x^{k+1}-x^k}.
    \end{equation*}
\end{enumerate}
These conditions imply $\sum_k\norm{x^{k+1}-x^k}<\infty$. Hence, $\{x^k\}$ has a single limit $x^*$. At this limit, $0\in\partial\Psi(x^*)$; under standard constraint qualifications, $x^*$ is a KKT point of Eq.~\eqref{eq:model}. If the KL exponent at $x^*$ is $1/2$, the convergence rate is linear.
\end{theorem}

The proof uses the standard KL descent argument for tame nonconvex optimization and nonconvex ADMM~\cite{attouch2013convergence,wang2019admm}. \hyperref[app:convergence]{Appendix A} outlines the steps. The curvature bound of tanh alone does not establish sufficient descent for an arbitrary two-penalty schedule, and the practical three-start HTT solver does not establish the exact-block assumption. The theorem applies to the exact-block algorithm under the stated conditions, not unconditionally to every numerical run.

\subsection{Complexity}

Let $D=\sum_v d_v$. Precomputing all $X_v^\top X_v$ costs $O(n^2D)$. Each iteration is dominated by $V$ dense $n\times n$ linear solves in the $Z$-updates, costing $O(Vn^3)$ time in a direct implementation. The reconstruction and error updates cost $O(n^2D)$ across views. After rotation and FFT, the tensor update performs $O(n)$ SVDs of $V\times n$ matrices. This cost is below the dense $Z$-solve cost when $V\ll n$. HTT itself evaluates $O(nV)$ scalar problems per iteration, each with a fixed iteration cap.

The main storage requirements are $O(Vn^2)$ for representations, auxiliary variables, and multipliers, plus $O(nD)$ for the data and cached cross-products. Thus the asymptotic orders are comparable to other full-representation t-SVD-based MVSC methods. The nonconvex scalar solver does not change the dominant big-$O$ order, but dense cubic solves and quadratic memory remain the practical bottleneck for large $n$.

\section{Experiments}
\label{sec:experiments}

\subsection{Datasets}

The evaluation uses 23 real benchmark datasets covering faces, objects, scenes, documents, handwritten digits, and other visual features. Sample sizes range from 169 to 10,800, view numbers from two to six, and class numbers from four to 100. Table~\ref{tab:datasets} lists the dimensions. All views of a dataset use the same sample ordering and cross-view correspondence.

\begin{table*}[t]
\centering
\caption{Summary of the 23 benchmark datasets. Entries in the last column follow the view order.}
\label{tab:datasets}
\scriptsize
\setlength{\tabcolsep}{4pt}
\begin{tabular}{lrrr p{6.4cm}}
\toprule
Dataset & Samples $n$ & Views $V$ & Classes $K$ & View dimensions \\
\midrule
Yale & 165 & 3 & 15 & 4096, 3304, 6750 \\
ORL & 400 & 3 & 40 & 4096, 3304, 6750 \\
COIL20 & 1440 & 4 & 20 & 1024, 944, 4096, 576 \\
Scene15 & 4485 & 3 & 15 & 1800, 1180, 1240 \\
MSRCv1 & 210 & 5 & 7 & 24, 576, 512, 256, 254 \\
Caltech101-7 & 1474 & 6 & 7 & 48, 40, 254, 1984, 512, 928 \\
EYaleB10 & 640 & 3 & 10 & 1024, 1239, 256 \\
BBC & 685 & 4 & 5 & 4659, 4633, 4665, 4684 \\
BBCSport & 544 & 2 & 5 & 3183, 3203 \\
NGs & 500 & 3 & 5 & 2000, 2000, 2000 \\
3Sources & 169 & 3 & 6 & 3560, 3631, 3068 \\
WebKB & 203 & 3 & 4 & 1703, 230, 230 \\
Handwritten & 2000 & 6 & 10 & 216, 76, 64, 6, 240, 47 \\
UCI & 2000 & 3 & 10 & 240, 76, 6 \\
HDigit & 10000 & 2 & 10 & 784, 256 \\
MNIST4 & 4000 & 3 & 4 & 30, 9, 30 \\
HW1256 & 2000 & 4 & 10 & 76, 216, 47, 6 \\
HW2Sources & 2000 & 2 & 10 & 784, 256 \\
100Leaves & 1600 & 3 & 100 & 64, 64, 64 \\
BDGP & 2500 & 4 & 5 & 1000, 500, 250, 79 \\
CCV & 6773 & 3 & 20 & 20, 20, 20 \\
ALOI100 & 10800 & 4 & 100 & 77, 13, 64, 125 \\
Scene & 2688 & 4 & 8 & 512, 432, 256, 48 \\
\bottomrule
\end{tabular}
\end{table*}

\subsection{Compared methods and implementation details}

The comparison uses nine representative methods. GMC~\cite{wang2020gmc} is the graph-based baseline. The others are LT-MSC, TLSpNM, LTBPL, LMVSC, LRTGL, t-SVD-MSC, GFT-TMVC, and GNLTA. They cover Tucker and t-SVD tensor models, nonconvex penalties, graph and proximity learning, graph-spectral transforms, and anchor-based clustering.

Each view uses the normalization routine supplied with its implementation. Baseline parameters follow the authors' recommended settings or dataset-specific code configurations. HTNN-MSC sets $\theta$ adaptively with $c=1$. The parameter studies consider
\begin{equation}
\lambda\in\{10^{-4},5\times10^{-4},10^{-3},5\times10^{-3},5\times10^{-2},0.5,5,10,100\}.
\end{equation}
For each dataset, the main tables report one post-processing run at the $\lambda$ with the best ACC. This oracle grid-selection protocol has no separate validation split and gives optimistic benchmark scores rather than held-out generalization estimates. Section~\ref{sec:repeatability} evaluates repeated post-processing at the selected parameters.

All algorithms run in MATLAB R2024a on a Windows workstation with 32~GB RAM using CPU computation. Running times cover the main optimization routines and exclude data loading and common spectral post-processing. All 230 method--dataset combinations completed. Our baseline adapters assign equal-weight feasible simplex solutions when GMC or LRTGL encounters zero denominators or empty graph rows. GMC also uses sparse partial eigendecomposition on CCV. On ALOI100, LT-MSC and GFT-TMVC use low-dimensional solves based on the Woodbury identity. Redundant memory is also cleared to fit the available RAM. These changes preserve their model objectives. Seven clustering metrics are computed: ACC, NMI, adjusted Rand index, F-score, precision, recall, and purity. The subsequent analysis focuses on ACC and NMI, which quantify label agreement and information consistency, respectively.

\subsection{Overall clustering performance}

% auto-generated by tools/make_latex_tables.py
\begin{table*}[t]
\centering
\caption{ACC results for ten methods across 23 datasets. \textbf{Bold} and \underline{underlined} values indicate the best and second-best results, respectively.}
\label{tab:acc}
\setlength{\tabcolsep}{3pt}
\scriptsize
\resizebox{\textwidth}{!}{%
\begin{tabular}{lcccccccccc}
\toprule
Dataset & LT-MSC & TLSpNM & LTBPL & GMC & LMVSC & LRTGL & t-SVD-MSC & GFT-TMVC & GNLTA & HTNN-MSC (ours) \\
\midrule
Yale & 0.7394 & \textbf{1.0000} & 0.7091 & 0.6545 & 0.5030 & 0.6727 & 0.8848 & \underline{0.9212} & 0.5885 & 0.9152 \\
ORL & 0.8175 & \underline{0.9350} & \textbf{0.9975} & 0.6325 & 0.4875 & 0.9225 & 0.8800 & 0.8000 & 0.8650 & 0.9250 \\
COIL20 & 0.7556 & 0.7736 & \underline{0.9250} & 0.8035 & 0.6472 & \textbf{0.9861} & 0.7542 & 0.7514 & 0.7581 & 0.7590 \\
Scene15 & 0.5663 & \textbf{0.8163} & 0.3278 & 0.3810 & 0.4970 & 0.6246 & 0.7897 & 0.5217 & 0.4211 & \underline{0.8096} \\
MSRCv1 & 0.8286 & \underline{0.9857} & \textbf{0.9905} & 0.7476 & 0.3429 & 0.7424 & 0.9524 & 0.8286 & 0.7405 & 0.8238 \\
Caltech101-7 & 0.6011 & 0.6282 & \underline{0.6649} & \textbf{0.6920} & 0.3365 & 0.6058 & 0.6567 & 0.5068 & 0.4094 & 0.6275 \\
EYaleB10 & 0.4672 & \textbf{0.7672} & 0.4203 & 0.5156 & 0.2781 & 0.5783 & 0.5766 & \underline{0.7125} & 0.2428 & 0.6234 \\
BBC & 0.5796 & 0.3285 & \textbf{1.0000} & 0.6934 & 0.4102 & 0.8934 & 0.3270 & 0.6102 & 0.9766 & \underline{0.9898} \\
BBCSport & 0.3971 & 0.3438 & \underline{0.9890} & 0.8070 & 0.3235 & 0.9449 & \textbf{0.9982} & 0.8474 & 0.9426 & 0.9651 \\
NGs & 0.9900 & 0.2020 & 0.0140 & 0.9820 & 0.2540 & 0.7560 & \textbf{1.0000} & 0.9780 & 0.6866 & \textbf{1.0000} \\
3Sources & \underline{0.7811} & 0.3491 & 0.0592 & 0.6923 & 0.4260 & \textbf{0.9467} & 0.3491 & 0.6864 & 0.2456 & 0.4320 \\
WebKB & 0.5369 & 0.5222 & 0.5419 & \textbf{0.7586} & 0.6404 & \underline{0.6946} & 0.5074 & 0.5517 & 0.5813 & 0.6256 \\
Handwritten & 0.9160 & \underline{0.9985} & \textbf{1.0000} & 0.8820 & 0.6550 & 0.9815 & 0.9975 & 0.9115 & 0.8890 & \underline{0.9985} \\
UCI & 0.8030 & \underline{0.9855} & \textbf{0.9990} & 0.7045 & 0.7135 & 0.9775 & 0.9650 & 0.9215 & 0.8718 & \underline{0.9855} \\
HDigit & 0.7826 & 0.6449 & \textbf{0.9998} & \underline{0.9981} & 0.5424 & 0.9840 & 0.6557 & 0.5998 & 0.7197 & 0.8332 \\
MNIST4 & 0.6442 & 0.6518 & 0.7440 & \underline{0.9203} & 0.7270 & 0.6498 & 0.6525 & 0.6375 & 0.8466 & \textbf{0.9227} \\
HW1256 & 0.8045 & \textbf{0.9975} & 0.8970 & 0.7590 & 0.7260 & 0.8825 & 0.9960 & 0.8935 & 0.8862 & \textbf{0.9975} \\
HW2Sources & 0.8965 & 0.8250 & 0.1025 & \underline{0.9940} & 0.5415 & 0.8785 & 0.8900 & 0.5930 & 0.7448 & \textbf{0.9995} \\
100Leaves & 0.7531 & 0.9056 & 0.9069 & 0.8237 & 0.3762 & 0.8647 & \underline{0.9200} & 0.6894 & 0.8166 & \textbf{0.9331} \\
BDGP & 0.5272 & 0.5248 & 0.2028 & 0.2036 & 0.4248 & \textbf{0.9410} & 0.5268 & 0.7116 & 0.7208 & \underline{0.9376} \\
CCV & 0.2153 & \underline{0.5368} & 0.1364 & 0.1160 & 0.1410 & 0.2026 & 0.4781 & 0.2101 & 0.1881 & \textbf{0.5649} \\
ALOI100 & 0.6018 & \textbf{0.7889} & 0.6312 & 0.7207 & 0.3925 & \underline{0.7825} & 0.7227 & 0.6972 & 0.6561 & 0.7663 \\
Scene & 0.7128 & \textbf{0.9003} & 0.5934 & 0.3400 & 0.5097 & 0.6670 & 0.8795 & 0.3586 & 0.4211 & \underline{0.8951} \\
\midrule
Mean & 0.6834 & 0.7135 & 0.6457 & 0.6879 & 0.4737 & 0.7904 & 0.7548 & 0.6930 & 0.6617 & 0.8404 \\
Avg. Rank & 6.07 & 4.39 & 5.09 & 5.96 & 8.57 & 4.43 & 4.70 & 6.28 & 6.74 & 2.78 \\
\bottomrule
\end{tabular}%
}
\end{table*}

% auto-generated by tools/make_latex_tables.py
\begin{table*}[t]
\centering
\caption{NMI results for ten methods across 23 datasets. \textbf{Bold} and \underline{underlined} values indicate the best and second-best results, respectively.}
\label{tab:nmi}
\setlength{\tabcolsep}{3pt}
\scriptsize
\resizebox{\textwidth}{!}{%
\begin{tabular}{lcccccccccc}
\toprule
Dataset & LT-MSC & TLSpNM & LTBPL & GMC & LMVSC & LRTGL & t-SVD-MSC & GFT-TMVC & GNLTA & HTNN-MSC (ours) \\
\midrule
Yale & 0.7674 & \textbf{1.0000} & 0.7959 & 0.6736 & 0.5323 & 0.7167 & 0.9017 & 0.9020 & 0.6441 & \underline{0.9538} \\
ORL & 0.9215 & \underline{0.9851} & \textbf{0.9977} & 0.8035 & 0.6848 & 0.9668 & 0.9641 & 0.9022 & 0.9673 & 0.9794 \\
COIL20 & 0.8369 & 0.8553 & \underline{0.9732} & 0.9176 & 0.7345 & \textbf{0.9866} & 0.8418 & 0.8267 & 0.8371 & 0.8303 \\
Scene15 & 0.5741 & \textbf{0.8680} & 0.3355 & 0.4155 & 0.4944 & 0.6619 & 0.8082 & 0.5263 & 0.4387 & \underline{0.8495} \\
MSRCv1 & 0.7421 & \underline{0.9758} & \textbf{0.9784} & 0.7421 & 0.2432 & 0.7160 & 0.9111 & 0.7396 & 0.7680 & 0.7907 \\
Caltech101-7 & 0.6380 & \textbf{0.7011} & 0.6457 & 0.6056 & 0.1246 & 0.5574 & \underline{0.6518} & 0.4771 & 0.4791 & 0.5968 \\
EYaleB10 & 0.4613 & \textbf{0.7772} & 0.3995 & 0.4929 & 0.2356 & 0.5741 & 0.5715 & \underline{0.7095} & 0.1878 & 0.6284 \\
BBC & 0.4479 & 0.0103 & \textbf{1.0000} & 0.4852 & 0.0829 & 0.7689 & 0.0073 & 0.4869 & 0.9371 & \underline{0.9651} \\
BBCSport & 0.1797 & 0.0145 & \underline{0.9612} & 0.7226 & 0.1150 & 0.8720 & \textbf{0.9938} & 0.6929 & 0.9421 & 0.9070 \\
NGs & 0.9652 & 0.0156 & 0.2587 & 0.9392 & 0.0738 & 0.8412 & \textbf{1.0000} & 0.9322 & 0.5849 & \textbf{1.0000} \\
3Sources & \underline{0.6998} & 0.0667 & 0.3155 & 0.5480 & 0.1789 & \textbf{0.8654} & 0.0535 & 0.6502 & 0.0452 & 0.2144 \\
WebKB & 0.1649 & 0.0363 & 0.0331 & \underline{0.3569} & 0.3386 & \textbf{0.4268} & 0.0157 & 0.1525 & 0.3400 & 0.2646 \\
Handwritten & 0.8519 & \underline{0.9962} & \textbf{1.0000} & 0.8932 & 0.6378 & 0.9544 & 0.9938 & 0.8369 & 0.8960 & 0.9962 \\
UCI & 0.7751 & 0.9653 & \textbf{0.9973} & 0.7548 & 0.7023 & 0.9460 & 0.9325 & 0.8469 & 0.8205 & \underline{0.9660} \\
HDigit & 0.6623 & 0.6093 & \textbf{0.9993} & \underline{0.9939} & 0.4812 & 0.9607 & 0.6154 & 0.5555 & 0.6671 & 0.7249 \\
MNIST4 & 0.6650 & 0.6751 & 0.6561 & \textbf{0.8071} & 0.5052 & 0.6785 & 0.6554 & 0.6045 & 0.6836 & \underline{0.8013} \\
HW1256 & 0.7849 & \textbf{0.9939} & 0.9318 & 0.8226 & 0.6660 & 0.8995 & 0.9908 & 0.8132 & 0.8384 & \underline{0.9939} \\
HW2Sources & 0.8105 & 0.7237 & 0.0045 & \underline{0.9853} & 0.4988 & 0.9413 & 0.7970 & 0.5569 & 0.6899 & \textbf{0.9986} \\
100Leaves & 0.8743 & \textbf{0.9814} & 0.9540 & 0.9025 & 0.6413 & 0.9417 & 0.9759 & 0.8426 & 0.9498 & \underline{0.9807} \\
BDGP & 0.3889 & 0.3821 & 0.0028 & 0.0044 & 0.2872 & \underline{0.8592} & 0.3898 & 0.6382 & 0.5184 & \textbf{0.8595} \\
CCV & 0.1746 & \underline{0.4805} & 0.0481 & 0.0167 & 0.0956 & 0.1738 & 0.4096 & 0.1716 & 0.1515 & \textbf{0.4952} \\
ALOI100 & 0.7557 & \textbf{0.8792} & 0.6703 & 0.7436 & 0.6140 & \underline{0.8772} & 0.8339 & 0.7925 & 0.8118 & 0.8728 \\
Scene & 0.5693 & \textbf{0.8082} & 0.5612 & 0.3142 & 0.4214 & 0.6115 & 0.7725 & 0.2692 & 0.3191 & \underline{0.7987} \\
\midrule
Mean & 0.6396 & 0.6435 & 0.6313 & 0.6496 & 0.4082 & 0.7738 & 0.6994 & 0.6490 & 0.6312 & 0.8029 \\
Avg. Rank & 6.09 & 4.09 & 5.09 & 6.04 & 8.91 & 4.22 & 4.76 & 6.74 & 6.04 & 3.02 \\
\bottomrule
\end{tabular}%
}
\end{table*}

Tables~\ref{tab:acc} and~\ref{tab:nmi} report ACC and NMI. HTNN-MSC achieves the highest mean ACC of 0.8404, compared with 0.7904 for LRTGL, and the best average ACC rank of 2.78. Its mean NMI is 0.8029 and its average NMI rank is 3.02. LRTGL records a mean NMI of 0.7738 and an average rank of 4.22. Several baselines achieve the best score on individual datasets, so the aggregate advantage does not amount to uniform dominance.

The nonconvex model improves substantially over the convex t-SVD-MSC skeleton on some text datasets. On BBC, HTNN-MSC attains an ACC of 0.9898, while t-SVD-MSC attains 0.3270. Both methods attain 1.0000 on NGs. BBCSport favors the convex skeleton, whose ACC is 0.9982 compared with 0.9651 for HTNN-MSC. On ALOI100, the largest dataset with 10,800 samples and 100 classes, HTNN-MSC attains an ACC of 0.7663 and an NMI of 0.8728. TLSpNM performs better with 0.7889 ACC and 0.8792 NMI.

Other datasets expose weaker cases for HTNN-MSC. Its ACC on 3Sources is 0.4320, whereas LRTGL and LT-MSC attain 0.9467 and 0.7811. HTNN-MSC also trails the best methods on MSRCv1 and Caltech101-7. Section~\ref{sec:discussion} examines these limitations.

\subsection{Cross-dataset statistical analysis}

% auto-generated by tools/make_statistical_tests.py
\begin{table*}[t]
\centering
\caption{Two-sided Wilcoxon signed-rank comparisons of HTNN-MSC against each baseline over 23 datasets. W/T/L counts higher, tied, and lower HTNN-MSC scores. The quantity $\Delta$ is the median paired difference. The value $p_{\mathrm{Holm}}$ adjusts nine comparisons separately for each metric. Bold values indicate $p_{\mathrm{Holm}}<0.05$.}
\label{tab:statistical_tests}
\setlength{\tabcolsep}{5pt}
\small
\begin{tabular}{lcccccc}
\toprule
& \multicolumn{3}{c}{ACC} & \multicolumn{3}{c}{NMI} \\
\cmidrule(lr){2-4}\cmidrule(lr){5-7}
Baseline & W/T/L & $\Delta$ & $p_{\mathrm{Holm}}$ & W/T/L & $\Delta$ & $p_{\mathrm{Holm}}$ \\
\midrule
LT-MSC & 21/0/2 & +0.1645 & \textbf{0.0007} & 20/0/3 & +0.1443 & \textbf{0.0015}\\
TLSpNM & 11/3/9 & +0.0000 & 0.0831 & 11/0/12 & +0.0000 & 0.2467\\
LTBPL & 14/0/9 & +0.1005 & \textbf{0.0446} & 13/0/10 & +0.0621 & 0.1688\\
GMC & 18/0/5 & +0.1094 & \textbf{0.0217} & 17/0/6 & +0.1292 & \textbf{0.0373}\\
LMVSC & 22/0/1 & +0.3738 & \textbf{<0.0001} & 22/0/1 & +0.3584 & \textbf{<0.0001}\\
LRTGL & 17/0/6 & +0.0452 & 0.0831 & 18/0/5 & +0.0418 & 0.1688\\
t-SVD-MSC & 19/1/3 & +0.0303 & \textbf{0.0165} & 18/1/4 & +0.0390 & \textbf{0.0373}\\
GFT-TMVC & 19/0/4 & +0.1176 & \textbf{0.0024} & 21/0/2 & +0.1381 & \textbf{0.0018}\\
GNLTA & 23/0/0 & +0.1137 & \textbf{<0.0001} & 20/0/3 & +0.1177 & \textbf{0.0003}\\
\bottomrule
\end{tabular}
\end{table*}

We follow the multiple-dataset comparison protocol in~\cite{demsar2006statistical}. Each dataset forms a block, tied methods receive average ranks, and the Friedman test is applied separately to ACC and NMI. For ACC, $\chi^2(9)=56.888$ and $p=5.31\times10^{-9}$. For NMI, $\chi^2(9)=61.790$ and $p=6.05\times10^{-10}$. Both tests reject the null hypothesis that all ten methods have equivalent performance rankings.

Two-sided Wilcoxon signed-rank tests compare HTNN-MSC with each baseline over the 23 paired dataset results. Each pairwise test omits zero differences. Holm's procedure controls the family-wise error rate across the nine comparisons for each metric~\cite{holm1979simple}. Table~\ref{tab:statistical_tests} gives the paired outcomes, median differences, and adjusted $p$-values. At the 0.05 level, seven ACC comparisons are significant. For TLSpNM and LRTGL, the ACC differences are nonsignificant; their Holm-adjusted $p$-values both equal 0.0831. Six NMI comparisons are significant. Nonsignificant NMI outcomes occur for TLSpNM ($p_{\mathrm{Holm}}=0.2467$) and for LTBPL and LRTGL (both $p_{\mathrm{Holm}}=0.1688$). The tests therefore detect many, but not all, pairwise differences. Since the scores come from oracle grid-selected parameters, they describe benchmark evidence rather than held-out generalization.

\subsection{Controlled comparison with convex soft thresholding}

We compare HTNN-MSC with t-SVD-MSC using the same self-expression skeleton and affinity post-processing to isolate the tensor penalty. HTNN-MSC has higher ACC on 19 datasets, equal ACC on one, and lower ACC on three. Mean ACC rises from 0.7548 to 0.8404, a gain of 8.56 percentage points. The gains on BBC, BDGP, MNIST4, and HDigit are 0.6628, 0.4108, 0.2702, and 0.1775. The losses on MSRCv1, BBCSport, and Caltech101-7 are 0.1286, 0.0331, and 0.0292.

This comparison measures the effect of replacing convex soft thresholding within a fixed skeleton. It cannot attribute every gain to asymptotic unbiasedness alone because the surrogate also changes the optimization path and effective regularization scale.

\subsection{Comparison among nonconvex surrogates}

% Exploratory same-skeleton surrogate comparison.
% Each entry is the best ACC over lambda in {5e-4, 5e-3, 0.5}; one ADMM run per cell.
\begin{table}[t]
\centering
\caption{Same-skeleton comparison of nonconvex rank surrogates. Each entry is the best ACC over $\lambda\in\{5\times10^{-4},5\times10^{-3},0.5\}$. These single-run results are exploratory. The best value in each row is bolded.}
\label{tab:surrogate_ablation}
\setlength{\tabcolsep}{5pt}
\small
\begin{tabular}{lcccc}
\toprule
Dataset & Tanh (ours) & Log & Schatten-$p$ & MCP \\
\midrule
Yale & 0.8485 & 0.9212 & 0.9152 & \textbf{0.9939} \\
ORL & 0.9675 & \textbf{0.9700} & 0.9650 & 0.9650 \\
MSRCv1 & \textbf{1.0000} & 0.9857 & 0.9857 & 0.9952 \\
BBCSport & 0.9651 & 0.9632 & \textbf{0.9963} & 0.9669 \\
BBC & \textbf{0.9898} & 0.9270 & 0.9737 & 0.9212 \\
Handwritten & \textbf{0.9995} & 0.9990 & 0.9990 & 0.9990 \\
\midrule
Mean ACC & 0.9617 & 0.9610 & 0.9725 & \textbf{0.9735} \\
Mean NMI & 0.9637 & 0.9446 & \textbf{0.9714} & 0.9526 \\
\bottomrule
\end{tabular}
\end{table}

A second controlled study replaces only the rank surrogate within the ADMM skeleton. We compare tanh, logarithmic, Schatten-$p$, and MCP penalties on Yale, ORL, MSRCv1, BBCSport, BBC, and Handwritten. Each surrogate is tested with $\lambda\in\{5\times10^{-4},5\times10^{-3},0.5\}$, and the table reports the best single-run ACC. The mean ACC values are 0.9617 for tanh, 0.9610 for log, 0.9725 for Schatten-$p$, and 0.9735 for MCP. Tanh achieves the best row value on three datasets but not the highest six-dataset average.

Tanh therefore does not uniformly outperform the other nonconvex penalties in this exploratory study. Its distinct features are boundedness, a Lipschitz derivative, the associated thresholding structure, and compatibility with an o-minimal KL analysis. Stronger rankings among surrogates would require repeated comparisons over a denser parameter grid.

\subsection{Parameter sensitivity}

\begin{figure}[t]
    \centering
    \includegraphics[width=\linewidth]{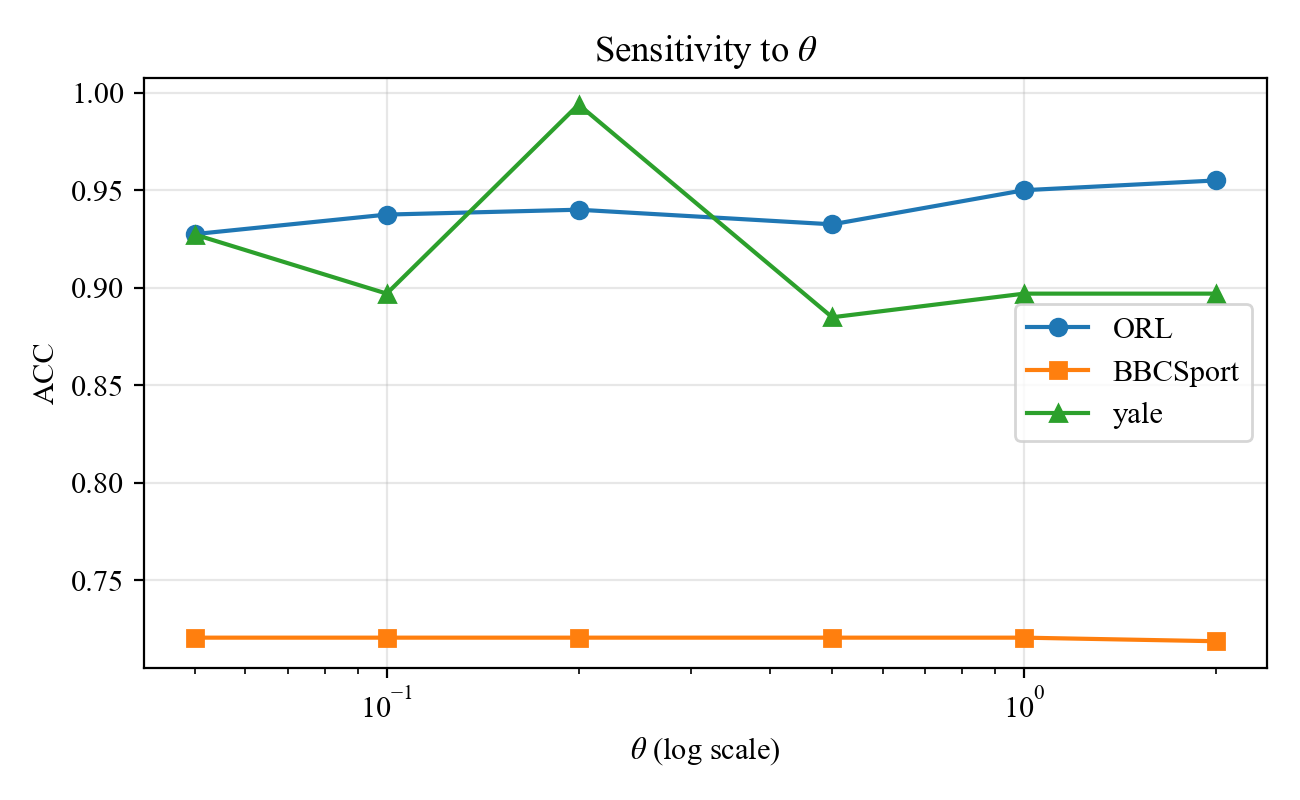}
    \caption{Sensitivity to the surrogate shape parameter. Adaptive scaling avoids choosing an absolute $\theta$ separately for every dataset, but does not guarantee the best point on each curve.}
    \label{fig:theta}
\end{figure}

Figure~\ref{fig:theta} shows the effect of fixed $\theta$ values on representative datasets. Stable regions coexist with dataset-dependent differences, so a common absolute scale is inappropriate. Adaptive scaling sets $\theta$ to the mean spectral magnitude and avoids dataset-specific $\theta$ tuning. This rule does not optimize $c$. On Yale, for example, the adaptive result remains below the best result in the broader fixed-shape exploration. Adaptive scaling is a practical default, not an optimality guarantee.

\begin{figure}[t]
    \centering
    \includegraphics[width=\linewidth]{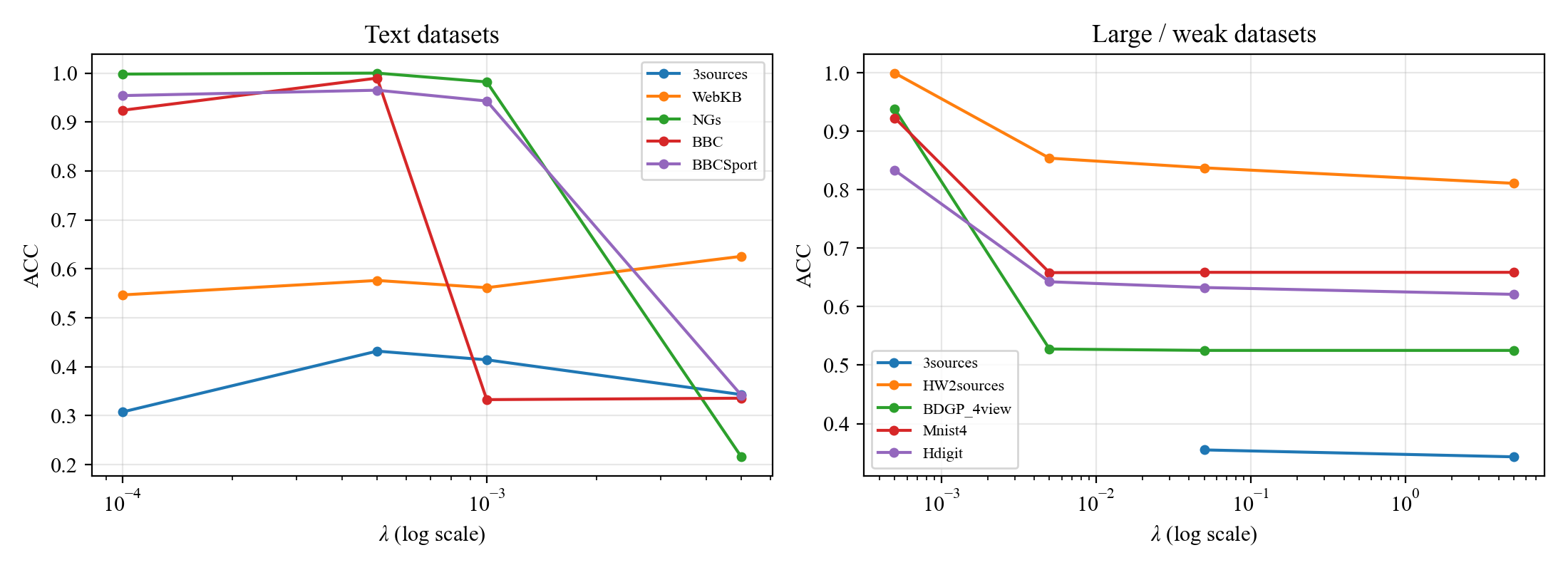}
    \caption{ACC sensitivity to $\lambda$ on representative datasets. The preferred range differs markedly between text and image or digit data.}
    \label{fig:lambda}
\end{figure}

The useful $\lambda$ range depends strongly on data type in Fig.~\ref{fig:lambda}. Several image and digit datasets favor values around 0.5 or above, whereas BBC, BBCSport, and NGs favor $5\times10^{-4}$. Excessive error weighting can substantially degrade the affinity. This three-order-of-magnitude variation explains why adaptive $\theta$ does not make HTNN-MSC parameter free.

\FloatBarrier
\subsection{Repeatability of spectral post-processing}
\label{sec:repeatability}

% auto-generated from our_method/exp8_repeat_corrected.csv
\begin{table}[H]
\centering
\caption{Repeatability of HTNN-MSC. For each dataset, ADMM runs once at the selected $\lambda$. Spectral post-processing is then repeated with ten fixed random seeds. Entries are mean $\pm$ sample standard deviation.}
\label{tab:repeatability}
\setlength{\tabcolsep}{4pt}
\scriptsize
\begin{tabular}{lccc}
\toprule
Dataset & $\lambda$ & ACC & NMI \\
\midrule
Yale & 0.5 & $0.9067\pm0.0268$ & $0.9483\pm0.0172$ \\
ORL & 0.5 & $0.9250\pm0.0000$ & $0.9794\pm0.0000$ \\
COIL20 & 10 & $0.7653\pm0.0000$ & $0.8354\pm0.0000$ \\
Scene15 & 0.5 & $0.8100\pm0.0000$ & $0.8497\pm0.0000$ \\
MSRCv1 & 10 & $0.8238\pm0.0000$ & $0.7907\pm0.0000$ \\
Caltech101-7 & 0.5 & $0.6275\pm0.0000$ & $0.5968\pm0.0000$ \\
EYaleB10 & 100 & $0.6234\pm0.0000$ & $0.6284\pm0.0000$ \\
BBC & 0.0005 & $0.9898\pm0.0000$ & $0.9651\pm0.0000$ \\
BBCSport & 0.0005 & $0.9651\pm0.0000$ & $0.9070\pm0.0000$ \\
NGs & 0.0005 & $1.0000\pm0.0000$ & $1.0000\pm0.0000$ \\
3Sources & 0.0005 & $0.4260\pm0.0000$ & $0.2171\pm0.0002$ \\
WebKB & 0.005 & $0.6256\pm0.0000$ & $0.2646\pm0.0000$ \\
Handwritten & 0.5 & $0.9985\pm0.0000$ & $0.9962\pm0.0000$ \\
UCI & 0.5 & $0.9855\pm0.0000$ & $0.9660\pm0.0000$ \\
HDigit & 0.0005 & $0.8334\pm0.0000$ & $0.7251\pm0.0000$ \\
MNIST4 & 0.0005 & $0.9227\pm0.0000$ & $0.8013\pm0.0000$ \\
HW1256 & 0.5 & $0.9980\pm0.0000$ & $0.9948\pm0.0000$ \\
HW2Sources & 0.0005 & $0.9995\pm0.0000$ & $0.9986\pm0.0000$ \\
100Leaves & 0.5 & $0.9206\pm0.0000$ & $0.9781\pm0.0000$ \\
BDGP & 0.0005 & $0.9376\pm0.0000$ & $0.8595\pm0.0000$ \\
CCV & 0.5 & $0.5647\pm0.0000$ & $0.4954\pm0.0000$ \\
ALOI100 & 0.5 & $0.7619\pm0.0000$ & $0.8694\pm0.0000$ \\
Scene & 0.5 & $0.8951\pm0.0000$ & $0.7987\pm0.0000$ \\
\bottomrule
\end{tabular}
\end{table}

The ADMM implementation is deterministic for fixed data and parameters. Randomness enters through $k$-means initialization in spectral post-processing. We run ADMM once at the selected $\lambda$, cache $S$, and repeat spectral clustering with ten fixed seeds. Table~\ref{tab:repeatability} reports the mean and sample standard deviation. Across the 23 datasets, the repeated runs have a mean ACC of 0.8394 and a mean NMI of 0.8029, close to the single-run summaries.

ACC variation is zero to numerical precision on 22 datasets. Yale has the only visible ACC variation, with ACC $0.9067\pm0.0268$ and NMI $0.9483\pm0.0172$. The NMI on 3Sources also varies slightly although its rounded ACC standard deviation is zero. Differences between single-run Yale results can thus arise from post-processing rather than different ADMM solutions. Small single-run margins require the same caution.

\FloatBarrier
\subsection{Convergence behavior and running time}

\begin{figure}[t]
    \centering
    \includegraphics[width=\linewidth]{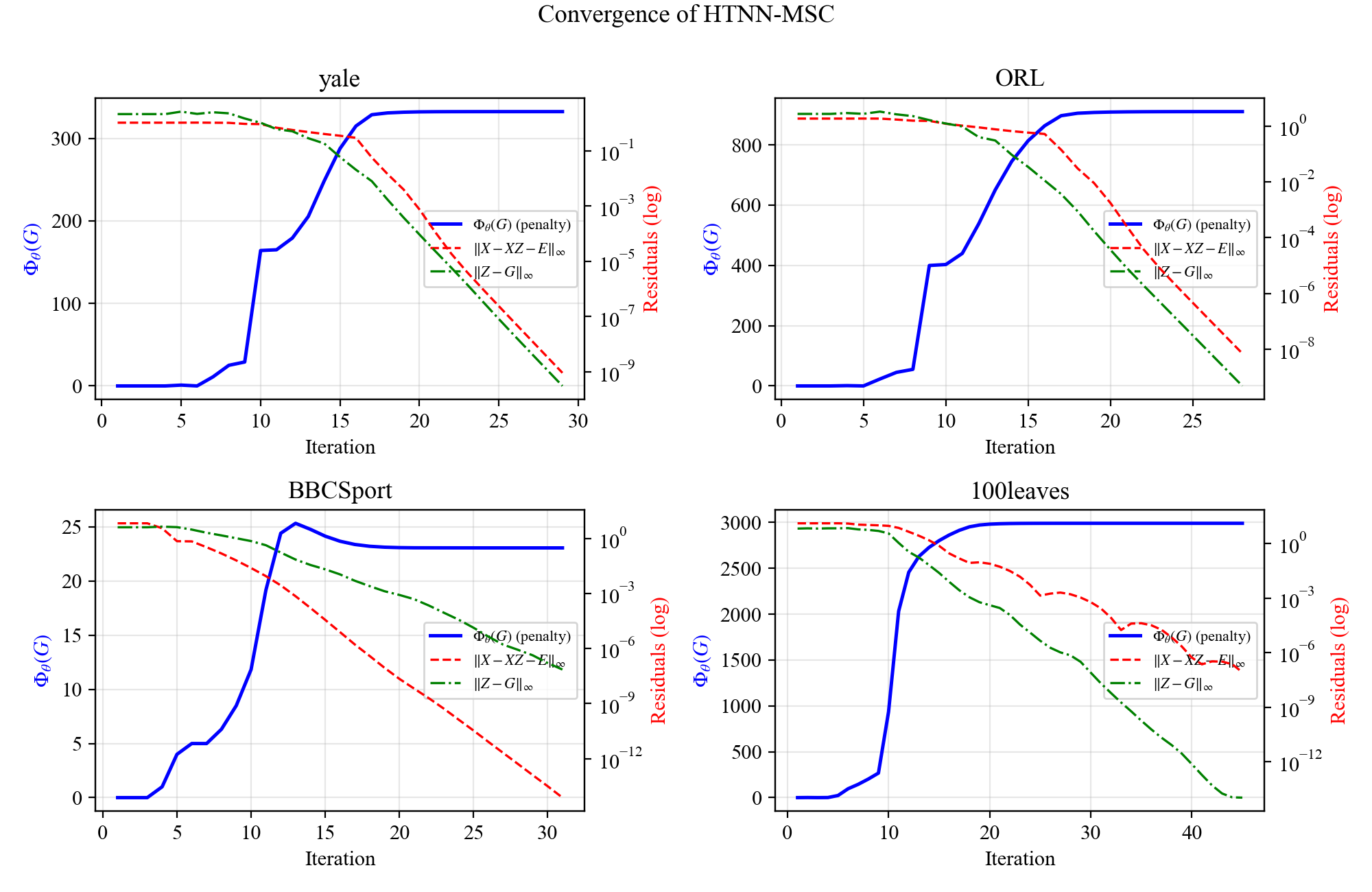}
    \caption{Empirical convergence of HTNN-MSC on four datasets. The plotted histories concern the numerical solver. Theorem~\ref{thm:convergence} additionally requires exact-block and descent conditions.}
    \label{fig:convergence}
\end{figure}

Figure~\ref{fig:convergence} plots reconstruction residuals $\max_v\norm{X_v-X_vZ_v-E_v}_\infty$, consistency residuals $\max_v\norm{Z_v-G_v}_\infty$, and the surrogate penalty on Yale, ORL, BBCSport, and 100Leaves. Both residual types decrease steadily in these runs and generally reach the stopping tolerance within approximately 30--45 iterations. The penalty grows from zero initialization and then stabilizes as the spectral structure forms.

\begin{figure}[t]
    \centering
    \includegraphics[width=\linewidth]{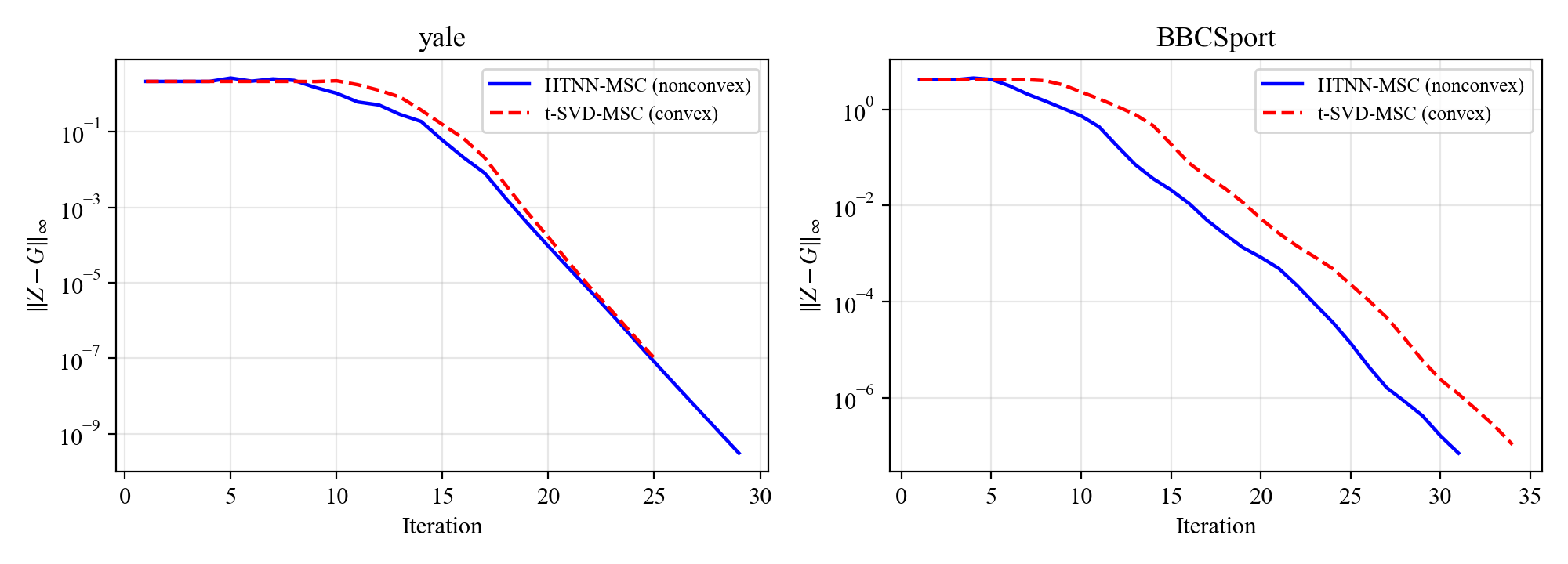}
    \caption{Residual histories for HTNN-MSC and the convex t-SVD-MSC skeleton on Yale and BBCSport.}
    \label{fig:convergence_compare}
\end{figure}

Figure~\ref{fig:convergence_compare} compares the HTNN and convex t-SVD skeletons on Yale and BBCSport. Their iteration counts are comparable, and the nonconvex version converges slightly faster in the displayed BBCSport run. These histories illustrate numerical behavior but do not replace the conditional convergence analysis.

% auto-generated: runtime comparison (seconds unless h)
\begin{table*}[t]
\centering
\caption{Running times in seconds, with h denoting hours.}
\label{tab:runtime}
\setlength{\tabcolsep}{3pt}
\scriptsize
\resizebox{\textwidth}{!}{%
\begin{tabular}{lcccccccccc}
\toprule
Dataset & LT-MSC & TLSpNM & LTBPL & GMC & LMVSC & LRTGL & t-SVD-MSC & GFT-TMVC & GNLTA & HTNN-MSC \\
\midrule
Yale & 16.1 & 6.1 & 1.8 & 0.3 & 41.5 & 12.7 & 3.9 & 6.6 & 0.4 & 16.4 \\
ORL & 35.1 & 14.2 & 3.8 & 0.5 & 0.2 & 39.3 & 11.2 & 15.9 & 1.0 & 24.6 \\
COIL20 & 299 & 106 & 91 & 6.0 & 0.3 & 261 & 73 & 54.4 & 16.3 & 96 \\
Scene15 & 1.7h & 697 & 2291 & 149 & 0.8 & 1.2h & 540 & 802 & 129 & 762 \\
MSRCv1 & 4.9 & 1.6 & 1.8 & 0.3 & 0.6 & 1.9 & 1.2 & 1.6 & 0.5 & 8.8 \\
Caltech101-7 & 389 & 99 & 102 & 7.3 & 0.2 & 321 & 70 & 126 & 25.7 & 121 \\
EYaleB10 & 30.8 & 8.8 & 15.3 & 1.1 & 0.1 & 13.2 & 8.8 & 16.1 & 2.3 & 27.0 \\
BBC & 131 & 40.5 & 18.2 & 24.7 & 0.5 & 101 & 37.2 & 49.7 & 3.9 & 90 \\
BBCSport & 26.1 & 11.1 & 7.8 & 6.5 & 0.6 & 28.7 & 10.8 & 11.4 & 1.3 & 28.1 \\
NGs & 26.8 & 9.5 & 11.9 & 0.6 & 0.3 & 15.2 & 9.5 & 9.7 & 1.4 & 10.5 \\
3Sources & 9.3 & 23.3 & 7.7 & 2.9 & 108 & 1615 & 4.2 & 4.5 & 0.4 & 8.8 \\
WebKB & 3.6 & 1.4 & 1.0 & 0.3 & 0.8 & 2.9 & 1.1 & 1.5 & 0.2 & 4.9 \\
Handwritten & 822 & 183 & 202 & 12.6 & 0.5 & 772 & 135 & 241 & 44.9 & 300 \\
UCI & 463 & 87 & 141 & 8.4 & 0.4 & 250 & 69 & 123 & 20.6 & 131 \\
HDigit & 7.6h & 1.1h & 3.0h & 482 & 1.5 & 9.0h & 2853 & 1.7h & 957 & 3201 \\
MNIST4 & 1.1h & 433 & 2097 & 104 & 0.6 & 1439 & 347 & 671 & 70 & 718 \\
HW1256 & 556 & 118 & 158 & 10.4 & 0.4 & 571 & 92 & 164 & 31.0 & 195 \\
HW2Sources & 390 & 57.9 & 161 & 6.4 & 0.3 & 339 & 43.9 & 64 & 12.9 & 58.2 \\
100Leaves & 207 & 60 & 65 & 4.3 & 0.9 & 220 & 40.0 & 76 & 14.4 & 68 \\
BDGP & 1139 & 205 & 251 & 10.6 & 0.3 & 1332 & 145 & 236 & 51.5 & 250 \\
CCV & 2.7h & 1860 & 2.4h & 93 & 1.6 & 1.0h & 863 & 2185 & 214 & 1484 \\
ALOI100 & 12.5h & 3.4h & 4.0h & 727 & 7.2 & 15.2h & 2.9h & 3.1h & 3375 & 2.7h \\
Scene & 1395 & 237 & 509 & 28.3 & 0.4 & 806 & 154 & 251 & 59.1 & 264 \\
\midrule
Total & 27.2h & 5.7h & 11.1h & 0.5h & 0.0h & 28.7h & 4.4h & 6.2h & 1.4h & 4.9h \\
\bottomrule
\end{tabular}%
}
\end{table*}

Table~\ref{tab:runtime} reports running times. HTNN-MSC spends approximately 4.9~h on the full benchmark, compared with 28.7~h for LRTGL and 27.2~h for LT-MSC. GMC, LMVSC, GNLTA, and t-SVD-MSC have lower total times. Large dense datasets remain expensive for HTNN-MSC, with HDigit and ALOI100 requiring about 0.89~h and 2.7~h. Implementation language, stopping behavior, and code optimization also affect these measurements. Runtime should therefore be interpreted together with asymptotic complexity rather than as a pure algorithmic ranking.

\FloatBarrier

\section{Discussion}
\label{sec:discussion}

HTNN-MSC achieves the best mean ACC and NMI, the best average ranks, and significant Friedman test results across the benchmark. It also records 19 wins, one tie, and three losses against the convex skeleton. These aggregates coexist with weaker results on individual datasets, nonsignificant Holm-adjusted pairwise comparisons, and better average ACC from other penalties in the surrogate ablation. The evidence supports the tested shrinkage strategy without implying dominance in every setting.

Five limitations delimit the current method. The best $\lambda$ varies by several orders of magnitude, and adaptive $\theta$ does not address error-weight selection. A label-free stability criterion could reduce this dependence. Results on 3Sources, MSRCv1, and Caltech101-7 also show that the tensor prior can be less effective than competing graph or tensor models. Heterogeneous view statistics, the sample-mode Fourier prior, and unsuitable $\lambda$ are possible factors, but the experiments do not isolate their effects. Oracle selection of $\lambda$ adds optimistic bias to the main benchmark. Repeated post-processing measures randomness only after selection, and the statistical tests do not correct this bias. Full-grid results, repeated complete pipelines, and label-free model selection would give a less optimistic evaluation. The $O(Vn^3)$ direct solves and $O(Vn^2)$ storage further restrict scalability. Anchor graphs, iterative linear solves, low-rank sketches, and graph-spectral transforms address different parts of this cost. The convergence theorem also assumes exact global block solutions and standard KL descent conditions. A certified bracketed HTT solver and a proof tailored to the implemented penalty schedule would connect the theorem more closely to practice.

\section{Conclusion}
\label{sec:conclusion}

The proposed HTNN-MSC incorporates a bounded hyperbolic tangent surrogate for tensor rank into MVSC. By allowing the marginal penalty to diminish with singular-value magnitude while remaining finite at the origin, the surrogate alleviates excessive asymptotic shrinkage of dominant components. Its two-penalty ADMM solver uses frequency-domain HTT and a data-adaptive surrogate scale. Under exact-block and descent assumptions, the o-minimal formulation yields a KL-based whole-sequence convergence result.

Across 23 datasets and nine baselines, HTNN-MSC achieves the best mean ACC and average ACC rank. A controlled comparison shows an aggregate gain over the convex t-SVD-MSC skeleton. The surrogate ablation also identifies cases where logarithmic, Schatten-$p$, or MCP penalties perform better. Repeatability and sensitivity experiments quantify spectral post-processing variation and the dependence on dataset-specific regularization. Certified scalar thresholding, label-free parameter selection, and scalable representations remain open tasks.

\appendix
\section{Additional convergence details}
\label{app:convergence}

\subsection{KL descent argument}

Under the sufficient-decrease condition in Theorem~\ref{thm:convergence}, $\Psi(x^k)$ is nonincreasing and $\norm{x^{k+1}-x^k}\to0$. Boundedness provides a convergent subsequence $x^{k_j}\to x^*$. Lower semicontinuity and the relative-error condition give $\Psi(x^{k_j})\to\Psi(x^*)$ and $0\in\partial\Psi(x^*)$ under the usual closedness assumptions.

The KL inequality at $x^*$ provides a concave desingularizing function $\varphi$ such that, on an appropriate neighborhood,
\begin{equation}
    \varphi'(\Psi(x)-\Psi(x^*))\,
    \dist(0,\partial\Psi(x))\geq1.
\end{equation}
Combining this inequality with sufficient decrease and relative error yields a bound on $\norm{x^{k+1}-x^k}$ in terms of successive differences of $\varphi(\Psi(x^k)-\Psi(x^*))$. Summability of the latter differences implies
\begin{equation}
    \sum_{k=0}^{\infty}\norm{x^{k+1}-x^k}<\infty.
\end{equation}
Therefore the whole sequence is Cauchy and converges to $x^*$. The standard recurrence for successive distances gives the linear-rate statement for KL exponent $1/2$.

This argument uses definability only to obtain the KL property. It does not replace the boundedness, descent, relative-error, or exact-block assumptions. In particular, a local curvature bound for tanh is insufficient by itself to establish those assumptions for arbitrary increasing $\mu_k$ and $\rho_k$.

\subsection{Certified scalar HTT bracketing}

An exhaustive HTT variant can bracket all stationary intervals. On $[0,t]$, global minimizers occur either at $s=0$ or at roots of $h$ in Eq.~\eqref{eq:stationary}. The derivative $h'$ has at most two zeros because the positive function $q(x)=\sech^2(x)\tanh(x)$ first increases and then decreases. These zeros partition $[0,t]$ into at most three intervals on which $h$ is monotone. Bisection on every interval whose endpoint values have opposite signs, together with objective comparison at $s=0$, returns the global minimizer up to the chosen root tolerance. This certified variant is a theoretical implementation option. The experiments use the three-start safeguarded Newton procedure described in Section~\ref{sec:optimization}.

\bibliographystyle{elsarticle-num}
\bibliography{references}

\section*{Declarations}

\subsection*{Funding}
Financial support from the University Development Program of Anhui Polytechnic University (HX-202112008, HX-202412103) and the Synergistic Innovation Program of Universities in Anhui Province (GXXT-2019-007) is gratefully acknowledged.

\subsection*{Conflict of interest}
The authors declare no competing financial or personal interests.

\subsection*{Author contributions (CRediT)}
\begingroup
\emergencystretch=2em
\textbf{Yiwen Dou:} Conceptualization, Funding acquisition, Investigation, Methodology, Project administration, Supervision, Writing---original draft, Writing---review \& editing. \textbf{Yiqiang Ding:} Data curation, Formal analysis, Software, Validation, Visualization.\par
\endgroup

\subsection*{Data availability}
Code and the reference scan module will be made publicly available upon acceptance.

\end{document}